\documentclass[10pt,twocolumn,letterpaper]{article}

\usepackage[pagenumbers]{iccv} 

\usepackage{amsmath}
\usepackage{graphicx}
\usepackage{algorithm}
\usepackage{algorithmic}
\usepackage{multirow}
\usepackage{graphicx}
\usepackage{amssymb}
\usepackage{float}

\def\para#1{\vspace{0.25em}\noindent\textbf{#1}}
\allowdisplaybreaks 

\definecolor{iccvblue}{rgb}{0.21,0.49,0.74}
\usepackage[pagebackref,breaklinks,colorlinks,allcolors=iccvblue]{hyperref}

\def\paperID{8867} 
\def\confName{ICCV}
\def\confYear{2025}

\title{LatentSync: Taming Audio-Conditioned Latent Diffusion Models for Lip Sync with SyncNet Supervision}

\author{
    \textbf{Chunyu Li}$^{1,2}$\qquad \textbf{Chao Zhang}$^1$\qquad \textbf{Weikai Xu}$^1$\qquad \textbf{Jingyu Lin}$^3$\qquad
\textbf{Jinghui Xie}$^{1,}$$^{\dag}$\\ \textbf{Weiguo Feng}$^1$\qquad \textbf{Bingyue Peng}$^1$\qquad \textbf{Cunjian Chen}$^3$\qquad \textbf{Weiwei Xing}$^{2,}$$^{\dag}$ \\[5pt] 
$^1$ByteDance \qquad $^2$Beijing Jiaotong University \qquad $^3$Monash University
}

\begin{document}
\maketitle
\begin{abstract}
    End-to-end audio-conditioned latent diffusion models (LDMs) have been widely adopted for audio-driven portrait animation, demonstrating their effectiveness in generating lifelike and high-resolution talking videos. However, direct application of audio-conditioned LDMs to lip-synchronization (lip-sync) tasks results in suboptimal lip-sync accuracy. Through an in-depth analysis, we identified the underlying cause as the ``shortcut learning problem'', wherein the model predominantly learns visual-visual shortcuts while neglecting the critical audio-visual correlations. To address this issue, we explored different approaches for integrating SyncNet supervision into audio-conditioned LDMs to explicitly enforce the learning of audio-visual correlations. Since the performance of SyncNet directly influences the lip-sync accuracy of the supervised model, the training of a well-converged SyncNet becomes crucial. We conducted the first comprehensive empirical studies to identify key factors affecting SyncNet convergence. Based on our analysis, we introduce StableSyncNet, with an architecture designed for stable convergence. Our StableSyncNet achieved a significant improvement in accuracy, increasing from 91\% to 94\% on the HDTF test set. Additionally, we introduce a novel Temporal Representation Alignment (TREPA) mechanism to enhance temporal consistency in the generated videos. Experimental results show that our method surpasses state-of-the-art lip-sync approaches across various evaluation metrics on the HDTF and VoxCeleb2 datasets. Code and models are publicly available at \url{https://github.com/bytedance/LatentSync}.
\end{abstract}

\section{Introduction}
\label{sec:intro}
The lip sync \cite{prajwal2020lip,mukhopadhyay2024diff2lip,guan2023stylesync,zhang2023dinet} is a video editing task, which regenerates the lip movements of a talking person according to the given audio, while maintaining the head pose and personal identity. This technique has broad applications in numerous practical domains, such as visual dubbing, virtual avatars, and video conferencing.

\begin{figure}[!t]
    \centering
    \includegraphics[width=\linewidth]{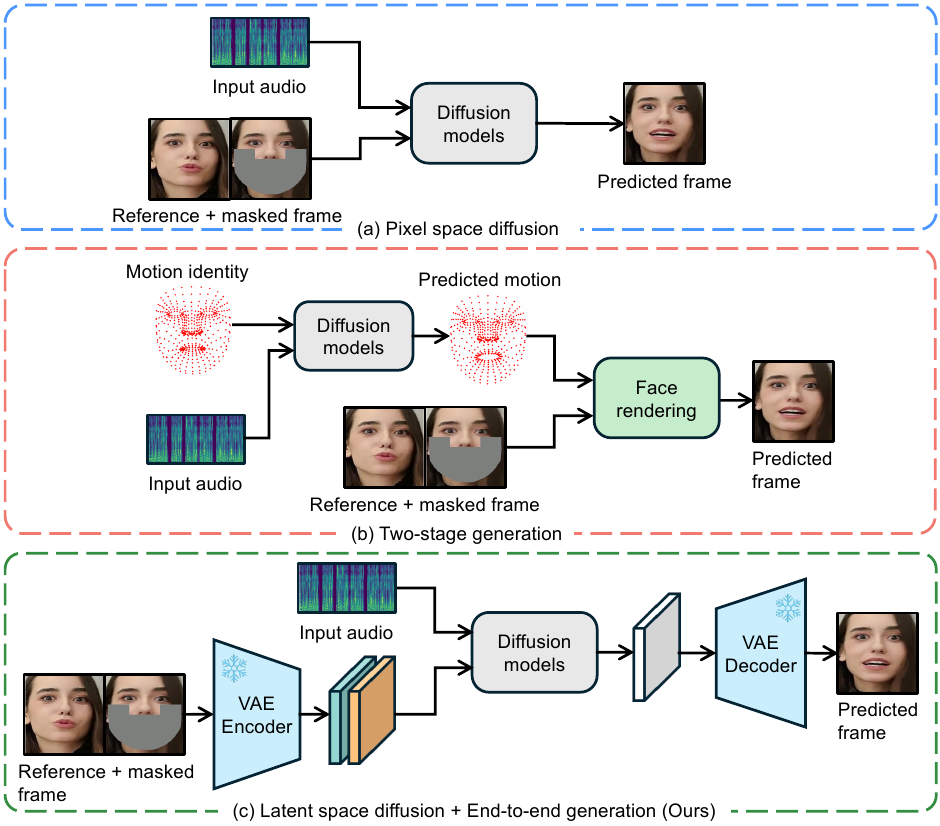}
    \caption{Frameworks comparison between previous diffusion-based lip-sync methods and our method.}
    \vspace{-10pt}
    \label{fig:frameworks_comparison}
\end{figure}

In the field of lip sync, GAN-based methods \cite{guan2023stylesync,prajwal2020lip} remain the mainstream approachs. The main issue with these methods is that they struggle to scale up \cite{kang2023scaling,sauer2023stylegan} to large and diverse datasets due to the unstable training \cite{arjovsky2017wasserstein,salimans2016improved} and mode collapse \cite{srivastava2017veegan,che2016mode}. Recent studies proposed diffusion-based methods \cite{mukhopadhyay2024diff2lip,bigioi2024speech,zhong2024style,yu2024make,liu2024diffdub} for lip sync, allowing the model to easily generalize across different individuals without the need for further fine-tuning on specific identities. However, these methods still have some limitations. Specifically, \cite{mukhopadhyay2024diff2lip,bigioi2024speech} perform the diffusion process in the pixel space (\cref{fig:frameworks_comparison} a), which restricts its ability to generate high-resolution videos due to the prohibitive hardware requirements. Other methods \cite{zhong2024style,yu2024make} adopt a two-stage approach: the first stage generates lip motions from audio, and the second stage synthesizes the visual appearance conditioned on the motion (\cref{fig:frameworks_comparison} b). The issue with this two-stage approach is that subtly different sounds may map to the same motion representation, leading to the loss of nuanced expressions linked to the emotional tone of the speech.

To address the above limitations, we propose \textit{LatentSync}, an end-to-end lip sync framework based on audio-conditioned LDMs \cite{rombach2022high} to generate lifelike and high-resolution talking videos, as shown in \cref{fig:frameworks_comparison} (c). We initially tried directly applying methods from the field of audio-driven portrait animation, such as EMO \cite{tian2024emo} and Hallo \cite{xu2024hallo}. However, the results showed poor lip-sync accuracy. We delved deeper into this phenomenon and identified the ``shortcut learning problem'' \cite{geirhos2020shortcut} inherent in lip sync task.

\para{The shortcut learning problem in lip sync.}
Lip-sync methods are typically based on a video-to-video inpainting framework, the model receives masked frames and audio as inputs. Unexpectedly, the audio-conditioned LDMs tends to predict lip movements based on visual information around the lips, such as facial muscles, eyes, and cheeks, while ignoring the audio information. We conducted an experiment to validate the existence of the shortcut learning problem and the effectiveness of SyncNet supervision \cite{prajwal2020lip} in mitigating this issue. We trained the audio-conditioned LDMs with masks of different sizes, both with and without SyncNet supervision. We used the sync confidence score \cite{chung2017out} to evaluate the synchronization accuracy between audio and lip movements.

\begin{figure}[h]
    \centering
    \includegraphics[width=0.8\linewidth]{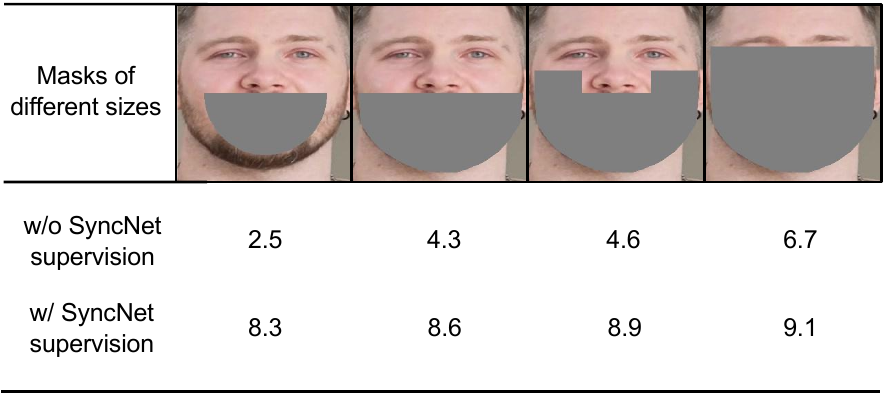}
    \caption{The shortcut learning problem in the lip-sync task. Higher sync confidence score means better lip-sync accuracy.}
    \vspace{-10pt}
    \label{fig:shortcut}
\end{figure}

As shown in \cref{fig:shortcut}, without SyncNet supervision, as the mask size increases, the visual information available in the masked frame for inferring lip movements decreases, forcing the model to rely more on audio information, thereby improving lip-sync accuracy, and vice versa. In contrast, with SyncNet supervision, the model remains focused on learning audio-visual correlations across different mask sizes.



\para{Why is SyncNet supervision not necessary for audio-driven portrait animation methods?}
These methods \cite{tian2024emo,xu2024hallo,chen2024echomimic,jiang2024loopy} are based on image-to-video framework. Since they do not involve input masked frames, they do not suffer from the shortcut learning problem.

The previous work Diff2Lip \cite{mukhopadhyay2024diff2lip} has explored how to add SyncNet supervision to pixel-space diffusion models. However, how to effectively apply SyncNet to latent diffusion models \cite{rombach2022high} remains unclear. Specifically, we explored two methods to incorporate SyncNet supervision into latent diffusion models: (a) Decoded pixel space supervision and (b) Latent space supervision. Furthermore, We found that SyncNet struggles to converge in both latent space and high-resolution pixel space. Since the convergence of SyncNet significantly impacts its supervision effectiveness, and ultimately affects the lip-sync accuracy of the supervised model, the convergence issue of SyncNet is highly valuable for research. Therefore, we conducted the first  comprehensive empirical studies in the aspects of model architecture design, training hyperparameters, and data preprocessing techniques, introducing the StableSyncNet with an architecture designed for stable convergence. Our StableSyncNet achieved the unprecedented 94\% accuracy on HDTF \cite{zhang2021flow}.

Additionally, we observed that high-frequency details in the generated talking videos, such as teeth, lips, and facial hair, exhibit flickering artifacts. We propose TREPA, a novel method designed to enhance temporal consistency and reduce such artifacts.


In summary, we made the following contributions: (1) We proposed LatentSync, the first lip-sync method that utilizes audio-conditioned LDMs to achieve end-to-end lifelike lip sync on high-resolution videos, incorporating TREPA to enhance the temporal consistency in the generated videos. (2) We identified the shortcut learning problem in lip-sync task and explored different methods to incorporate SyncNet supervision into audio-condtioned LDMs. (3) We conducted the first comprehensive empirical studies to identify key factors affecting SyncNet convergence, introducing the StableSyncNet for stable convergence.

\begin{figure*}[h]
    \centering
    \includegraphics[width=\linewidth]{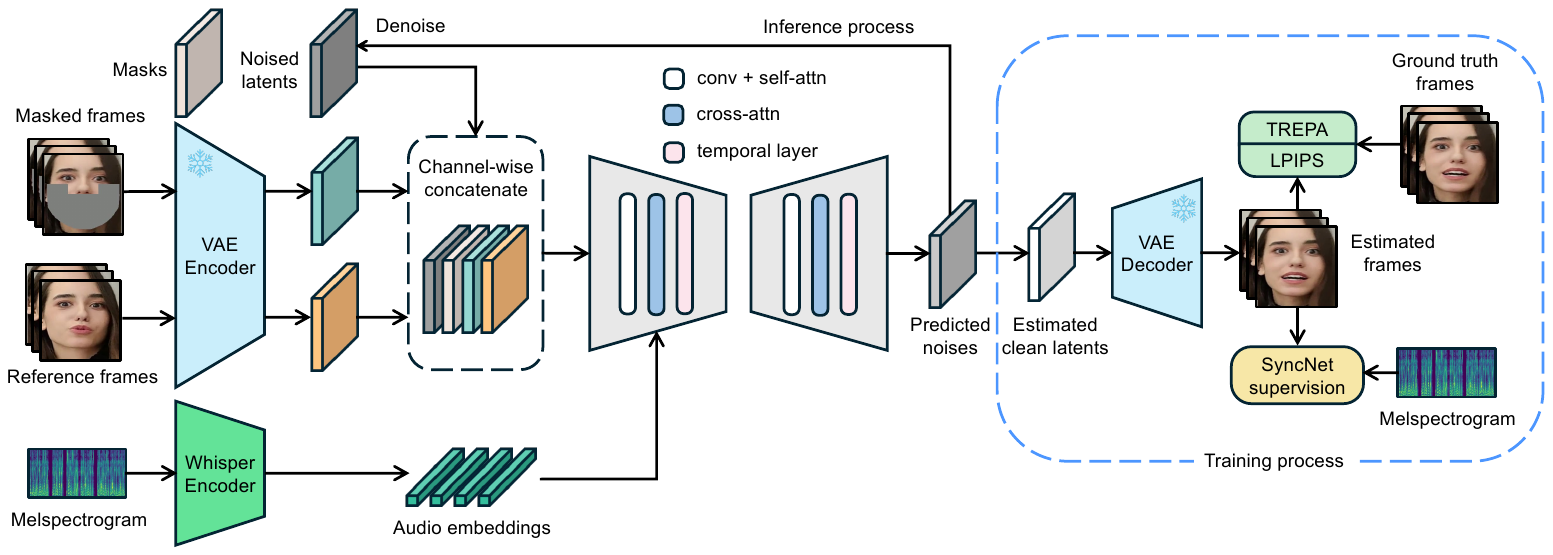}
    \caption{The overview of our LatentSync framework. We use the Whisper \cite{radford2023robust} to convert melspectrogram into audio embeddings, which are then integrated into the U-Net \cite{ronneberger2015u} via cross-attention layers. The reference and masked frames are channel-wise concatenated with noised latents as the input of U-Net. In the training process, we use a one-step method to get estimated clean latents from predicted noises, which are then decoded to obtain the estimated clean frames. The TREPA, LPIPS \cite{zhang2018unreasonable} and SyncNet loss \cite{prajwal2020lip} are added in the pixel space.}
    \vspace{-10pt}
    \label{fig:framework}
\end{figure*}
\section{Related Work}
\label{sec:related_work}

\subsection{Diffusion-based Lip Sync}
Diff2Lip \cite{mukhopadhyay2024diff2lip} and DrivenVideoEditing \cite{bigioi2024speech} are both end-to-end lip sync methods based on pixel-space audio-conditioned diffusion models. MyTalk \cite{yu2024make} uses diffusion models in the first stage to complete audio-to-motion conversion and uses a VAE \cite{kingma2013auto} in the second stage for motion-to-image generation. StyleSync \cite{zhong2024style} uses transformers in the first stage to convert audio to motion and employs diffusion models in the second stage for motion-to-image generation. DiffDub \cite{liu2024diffdub} uses diffusion autoencoders \cite{preechakul2022diffusion} to convert the masked images into semantic latent codes in the first stage, and uses diffusion models to generate an image conditioned on the semantic latent codes and audio in the second stage.

\subsection{Non-diffusion-based Lip Sync}
Wav2Lip \cite{prajwal2020lip} is the most classic lip sync method that introduced using a pretrained SyncNet \cite{chung2017out} to supervise the training of the lip-sync generator. \cite{gupta2023towards} trains a VQ-VAE \cite{esser2021taming} to encode faces and head poses, and then trains the lip sync generator in the quantized space to generate high-resolution images. StyleSync \cite{guan2023stylesync} follows the overall framework of Wav2Lip, with its main innovation being the use of StyleGAN2 \cite{karras2020analyzing} as the generator backbone. VideoReTalking \cite{cheng2022videoretalking} divides lip sync into three components: semantic-guided reenactment network, lip sync network, and identity-aware refinement and enhancement. DINet \cite{zhang2023dinet} deforms the feature maps to generate mouth shapes conditioned on the driving audio. MuseTalk \cite{zhang2024musetalk} uses the architecture of Stable Diffusion \cite{rombach2022high} for inpainting, but it does not perform diffusion process and uses a discriminator for adversarial learning \cite{goodfellow2014generative}, making it more like a GAN-based framework.

\subsection{Audio-driven Portrait Animation}
Many people may confuse lip sync with audio-driven portrait animation. These two tasks have some similarities but are actually completely different tasks. Lip sync is based on video-to-video editing framework, which needs to keep areas other than the mouth the same as in the input video. Audio-driven portrait animation is based on image-to-video animation framework, which can change the head movement and even facial expressions. Although there are already some audio-driven portrait animation methods based on audio-conditioned LDMs, such as \cite{tian2024emo,chen2024echomimic,xu2024hallo,jiang2024loopy}, they cannot be directly applied to the lip sync task due to the shortcut learning problem.
\section{Method}

\subsection{LatentSync Framework}
The overview of the LatentSync framework is shown in \cref{fig:framework}. The framework is based on video-to-video inpainting with the temporal modeling by temporal layer \cite{guo2023animatediff}. To incorporate the visual features of the face from the input video, reference frames are introduced as additional inputs. During training, these frames are randomly selected, while during inference, they are taken from the current frames. We concatenate different inputs along the channel dimension, making the total input of U-Net to be 13 channels (4 channels for noise latent, 1 channel for the mask, 4 channels for the masked frame, and 4 channels for the reference frame). At the beginning of training, the model is initialized with the parameters of SD 1.5 \cite{rombach2022high}, except for the first \texttt{conv\_in} layer with 13 channels and cross-attention layers of dimension 384, which are randomly initialized.

\para{Audio layers.}
We used the pretrained audio feature extractor Whisper \cite{radford2023robust} to extract audio embeddings. Lip motion may be influenced by the audio from the surrounding frames, and a larger range of audio input also provides more temporal information for the model. Therefore, for each generated frame, we bundled the audio from several surrounding frames as input. We define the input audio feature $A^{(f)}$ for the $f$th frame as: $A^{(f)} = \left\{ a^{(f-m)}, \dots, a^{(f)}, \dots, a^{(f+m)} \right\}$, where $m$ is the number of surrounding audio features from one side. To integrate the audio embeddings into the U-Net \cite{ronneberger2015u}, we used the native cross-attention layer.

\begin{figure}[h]
    \centering
    \includegraphics[width=0.7\linewidth]{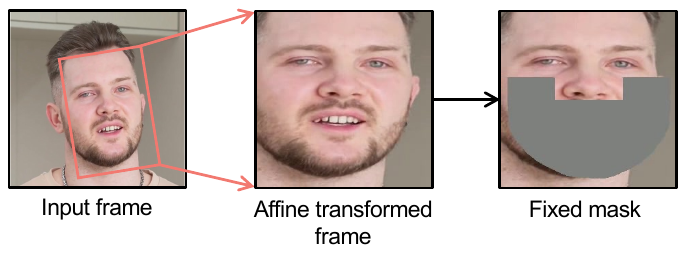}
    \caption{The illustration of affine transformation and fixed mask.}
    \vspace{-10pt}
    \label{fig:affine}
\end{figure}

\para{Affine transformation and fixed mask.}
During the data preprocessing stage, affine transformation was employed to perform face frontalization. This approach \cite{guan2023stylesync} helps the model to effectively learn facial features particularly in challenging scenarios such as side-profile views. We applied a mask that covers the entire face to minimize the model's tendency to learn visual-visual shortcuts. The position and shape of the mask are fixed. We do not use the detected landmarks \cite{lugaresi2019mediapipe,bulat2017far} to draw the mask, as moving landmarks also provide cues about lip movements. The affine transformation and fixed mask are illustrated in \cref{fig:affine}.

\begin{figure}[!t]
    \centering
    \includegraphics[width=0.8\linewidth]{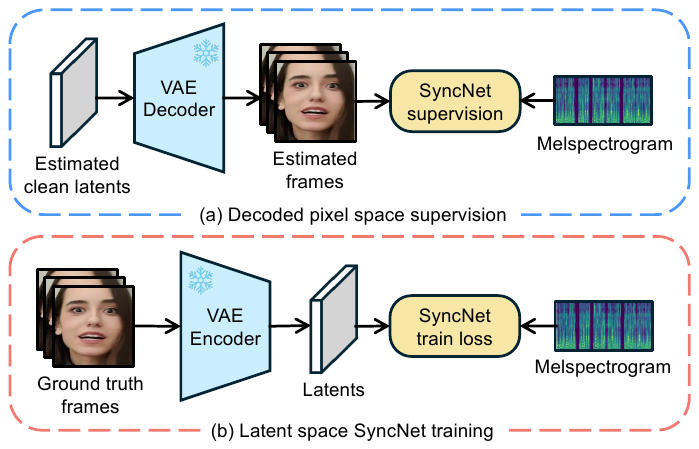}
    \caption{Two methods to add SyncNet supervision to latent diffusion models.}
    \vspace{-10pt}
    \label{fig:supervision}
\end{figure}

\para{SyncNet supervision.}
Latent diffusion models predict in the noise space, while SyncNet \cite{chung2017out,prajwal2020lip} requires an input in the image space. To address this issue, we use the predicted noise $\epsilon_\theta(z_t)$ to obtain the estimated $\hat{z}_0$ in one step, which can be formulated as:
\begin{equation}
    \hat{z}_0 = \left( z_t - \sqrt{1 - \bar{\alpha}_t} \, \epsilon_\theta(z_t) \right) / \sqrt{\bar{\alpha}_t}
\end{equation}
Another problem is that latent diffusion models make predictions in the latent space. We explored two methods to incorporate SyncNet supervision into latent diffusion models: (a) \textbf{Decoded pixel space supervision}, which trains SyncNet in the same way as Wav2Lip \cite{prajwal2020lip}. (b) \textbf{Latent space supervision}, which requires training a SyncNet in the latent space. The visual encoder input of this SyncNet is the latent vectors obtained by the VAE \cite{esser2021taming,kingma2013auto} encoding.  The illustration is shown in \cref{fig:supervision}. Our empirical analysis in \cref{sec:ablation} reveals that training SyncNet in the latent space exhibits inferior convergence compared to training in the pixel space. This degradation may arise from information loss in the lip region during the VAE encoding process. The poorer convergence of SyncNet in the latent space adversely impacts the lip-sync accuracy of the supervised diffusion models, according to the experimental results in \cref{sec:ablation}. Therefore, we finally choose the decoded pixel space supervision in the LatentSync framework.

\subsection{Two-Stage Training Strategy}

The decoded pixel space supervision has a problem that the activations in VAE decoding need to be stored for backpropagation, which significantly increases GPU memory consumption. To mitigate this issue, we designed a two-stage training strategy: in the first stage, the model learns to extract features from reference frames and develops inpainting capabilities, which we refer to as learning visual features. In the second stage, it learns audio-visual correlations under SyncNet supervision. This approach eliminates the need for the VAE decoding process in the first stage, enabling model to learn visual features with a larger batch size. We observed that audio-conditioned LDMs typically spend more time learning visual features and less time learning audio-visual correlations. Therefore, this two-stage training strategy allows the model to efficiently learn both visual features and audio-visual correlations. We provide the formal definitions of training objectives in the following paragraphs.

In the first stage of training, we do not add the temporal layer \cite{guo2023animatediff,blattmann2023align} and train all parameters of the U-Net. The training objective has only a simple loss \cite{ho2020denoising}:
\begin{equation}
    \mathcal{L}_{\text{simple}} = \mathbb{E}_{x, A, \epsilon \sim \mathcal{N}(0,1), t}
    \left[ \left\| \epsilon - \epsilon_\theta(z_t, t, \tau_\theta(A)) \right\|_2^2 \right]
\end{equation}
where $A$ is the input audio, $\epsilon_\theta(z_t, t, \tau_\theta(A))$ is the predicted noise, and $\tau_\theta$ is the audio feature extractor.

In the second stage of training, we only train the temporal layer and audio layer while freezing the other parameters of the U-Net. Suppose we have 16 decoded video frames $\mathcal{D}(\hat{z}_0)_{f:f+16}$ and the corresponding audio sequence $a_{f:f+16}$, the SyncNet loss can be formulated as:
\begin{equation}
    \mathcal{L}_{\text{sync}} = \mathbb{E}_{x, a, \epsilon, t} \left[ \text{SyncNet}(\mathcal{D}(\hat{z}_0)_{f:f+16}, a_{f:f+16}) \right]
\end{equation}
where $\mathcal{D}$ represents the VAE decoder, since the lipsync task requires generating detailed areas, such as lips, teeth, and facial hair, we used the LPIPS \cite{zhang2018unreasonable} to improve the visual quality of the images generated by the U-Net.
\begin{equation}
    \mathcal{L}_{\text{lpips}} = \mathbb{E}_{x, \epsilon, t}
    \left[ \left\| \mathcal{V}_l(\mathcal{D}(\hat{z}_0)_f) - \mathcal{V}_l(x_f) \right\|_2^2 \right]
\end{equation}
where $\mathcal{V}_l(\cdot)$ denotes the features extracted from the $l^{\text{th}}$ layer of a pretrained VGG network \cite{simonyan2014very}. In addition, to improve temporal consistency, we also employed the proposed TREPA, see \cref{eq:trepa}. For more details of the TREPA, please refer to \cref{sec:trepa}.

The total loss function for the second stage of training is:
\begin{equation}
    \mathcal{L}_{\text{total}} = \lambda_{1} \mathcal{L}_{\text{simple}} + \lambda_{2} \mathcal{L}_{\text{sync}} + \lambda_{3} \mathcal{L}_{\text{lpips}} + \lambda_{4} \mathcal{L}_{\text{trepa}}
\end{equation}


\subsection{Temporal Representation Alignment}
\label{sec:trepa}
TREPA aligns the temporal representations of the generated image sequences with those of ground truth image sequences. The insight behind this method is that merely employing distance loss between individual images improves the content quality of single generated images but does not enhance the temporal consistency of the generated image sequence. In contrast, temporal representations capture temporal correlation within image sequences, enabling the model to focus on improving the overall temporal consistency. We employed a large-scale self-supervised video model VideoMAE-v2 \cite{wang2023videomae} to extract temporal representations. Due to its unsupervised training on large-scale unlabeled datasets, the model's temporal representations exhibit strong generalization capabilities, robustness, and high information density.

In mathematical form, let $\mathcal{T}$ be the self-supervised video model encoder. The encoder's output is the embedding before the head projection. TREPA can be represented as:
\begin{equation}
    \mathcal{L}_{\text{trepa}} = \mathbb{E}_{x, \epsilon, t} \left[ \left\| \mathcal{T}(\mathcal{D}(\hat{z}_0)_{f:f+16}) - \mathcal{T}(x_{f:f+16}) \right\|_2^2 \right]
    \label{eq:trepa}
\end{equation}
where the straightforward Mean Squared Error (MSE) is employed to measure the distance between temporal representations. We also fix the representation by $\ell_2$ normalization before calculating the MSE.

\section{Empirical Studies on SyncNet Convergence}
\label{sec:syncnet}
Our experiments reveal that SyncNet is difficult to converge in both latent space and high-resolution pixel space, a typical characteristic of this issue is that the training loss gets stuck at 0.69 and fails to decrease further. In this section, we analyze the SyncNet convergence problem and identify several critical factors affecting the convergence through various ablation studies. Importantly, we preserved SyncNet's original training framework and employed the same contrastive loss function as Wav2Lip \cite{prajwal2020lip}. Therefore, our experience can be applied to many lip-sync \cite{prajwal2020lip,guan2023stylesync,mukhopadhyay2024diff2lip,cheng2022videoretalking,zhang2023dinet,zhang2024musetalk} and audio-driven portrait animation methods \cite{ma2023dreamtalk,ye2023geneface} that utilize SyncNet.

\para{Why is the SyncNet training loss stuck at 0.69?}
According to the classic training framework of SyncNet \cite{prajwal2020lip,chung2017out}, we randomly provide SyncNet with positive and negative samples with a 50\% probability during the training stage. The output of SyncNet is a probability distribution of whether the sample is positive or negative. We define $p(x=1)$ as the probability that the sample is positive, and $p(x=0)$ as the probability that the sample is negative. Let $p(x)$ be the true probability distribution and $q(x)$ the predicted probability distribution. When $q(x=1) \approx q(x=0) \approx 0.5$ and batch size $N$ is sufficiently large, the training loss of SyncNet is (proof in \cref{suppsec:syncnet_loss_proof}):
\begin{align}
    \mathcal{L}_{\text{syncnet}} & = -\frac{1}{N}\sum_{i=1}^N \sum_{x_i \in \{0,1\}} p(x_i) \,\text{log}\, {q(x_i)} \approx 0.693
\end{align}
This means that SyncNet has not learned any discriminative capability; it is just randomly guessing whether the samples are positive or negative. This may be due to various reasons, including the model's insufficient capacity to fit the data, the large audio-visual offset in the data, and flaws in the training strategy. In the following paragraphs, we will identify the key factors affecting the convergence of SyncNet through comprehensive ablation studies.

\begin{figure}[!t]
    \centering
    \includegraphics[width=0.9\linewidth]{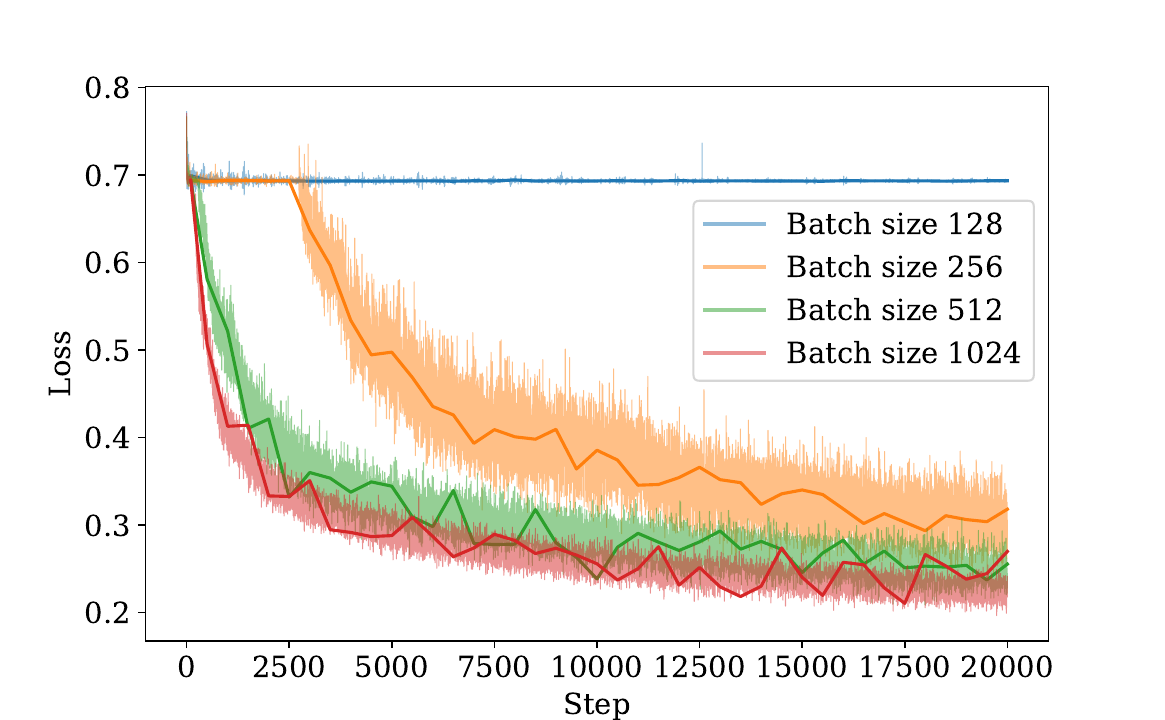}
    \caption{SyncNet training curves of different batch sizes. VoxCeleb2 results, the more transparent curves represent the training set loss, while the darker curves represent the validation set loss; the same applies to the following figures. (VoxCeleb2, Dim 2048, StbleSyncNet arch, 16 frames.)}
    \vspace{-10pt}
    \label{fig:batch_ablation}
\end{figure}

\para{Batch size.}
As shown in \cref{fig:batch_ablation}, a larger batch size (e.g., 1024) not only enables the model to converge faster and more stably but also results in a lower validation loss at the end of training. In contrast, smaller batch sizes (e.g., 128) may fail to converge, with the loss remaining stuck at 0.69. Even with a slightly larger batch size (e.g., 256), while convergence may be achieved, the training loss exhibits significant oscillations during its descent.

\begin{figure}[!t]
    \centering
    \includegraphics[width=0.9\linewidth]{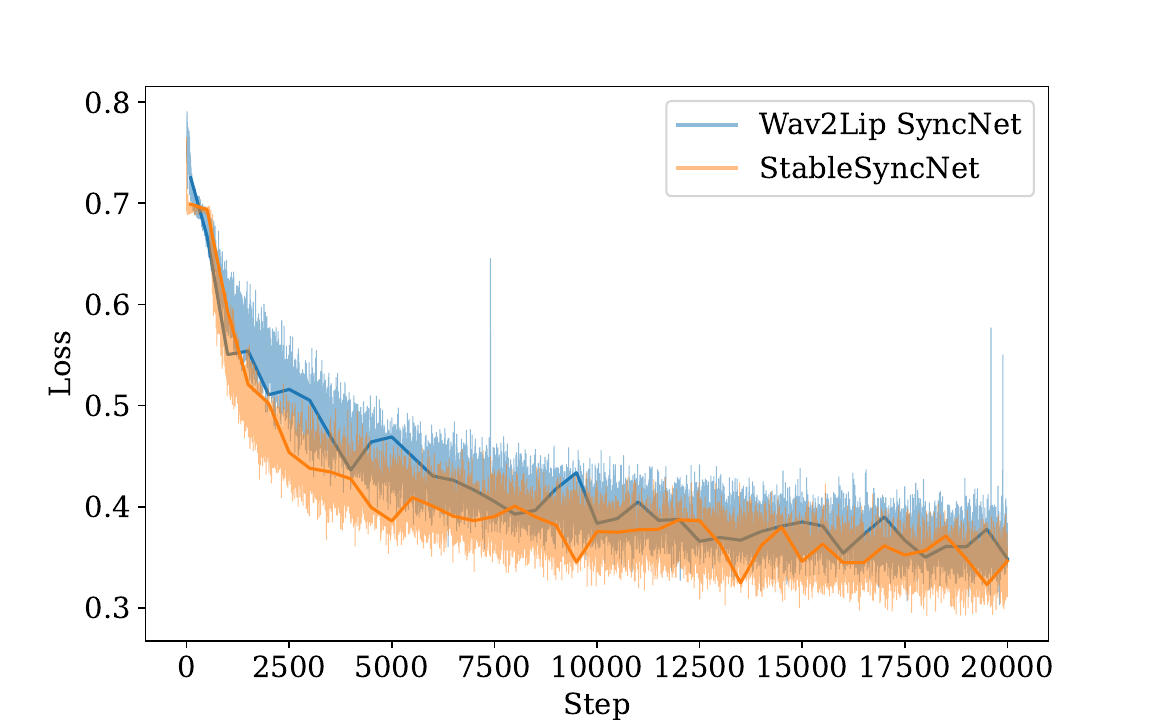}
    \caption{SyncNet training curves of different architectures. For comparison, we also modified the architecture of Wav2Lip's SyncNet to accept 256 $\times$ 256 visual input according to \cite{wav2lip288}. (VoxCeleb2, Dim 2048, Batch size 512, 5 frames.)}
    \vspace{-10pt}
    \label{fig:arch_ablation}
\end{figure}

\para{Architecture.}
We redesign the SyncNet's visual and audio encoders with the U-Net encoder from Stable Diffusion 1.5 \cite{rombach2022high}, retaining the structure of residual blocks \cite{he2016deep} and self-attention blocks \cite{vaswani2017attention} in the U-Net encoder blocks. We only adjusted the downsampling factors based on the size of the input visual images and mel-spectrograms, and we removed the cross-attention blocks, as SyncNet does not require additional conditions. We refer to the SyncNet with this modified architecture as \textit{StableSyncNet}. As shown in \cref{fig:arch_ablation}, StableSyncNet maintained both training loss and validation loss lower than those of Wav2Lip's SyncNet \cite{prajwal2020lip} throughout the training process.

\begin{figure}[!t]
    \centering
    \includegraphics[width=0.9\linewidth]{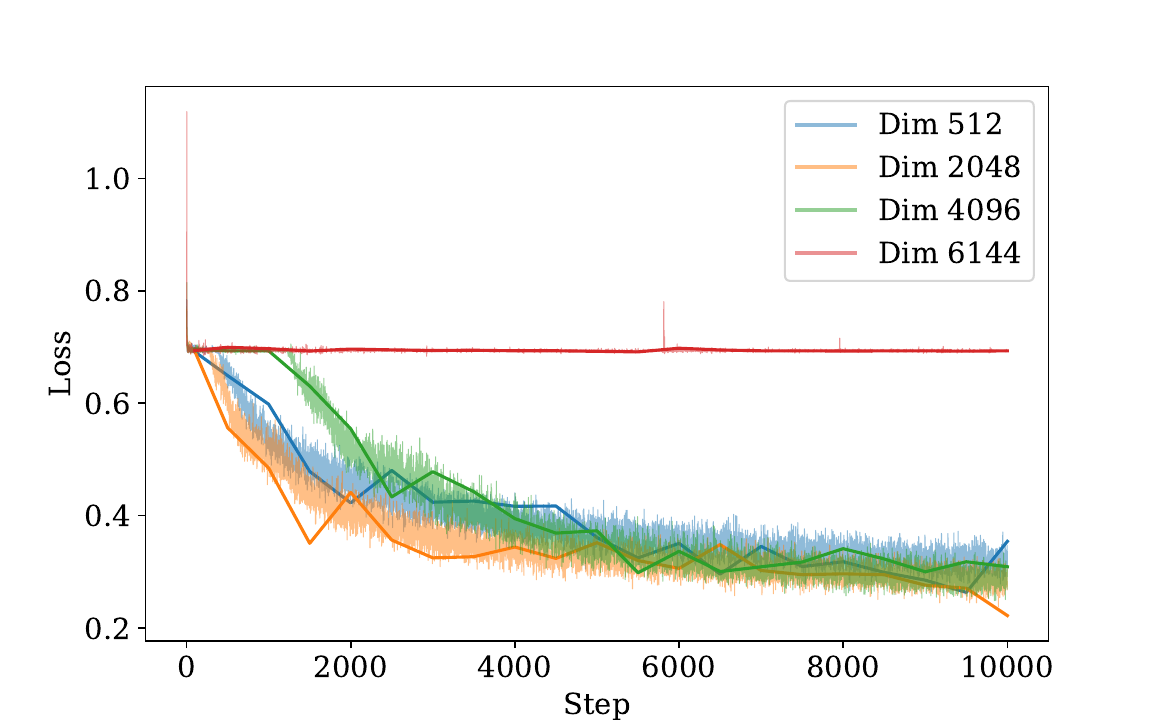}
    \caption{SyncNet training curves of different embedding dimensions. (VoxCeleb2, Batch size 512, StableSyncNet arch, 16 frames.)}
    \vspace{-10pt}
    \label{fig:dim_ablation}
\end{figure}

\para{Embedding dimension.}
As illustrated in \cref{fig:dim_ablation}, embeddings with smaller dimensions (e.g., 512) result in representations that fail to capture sufficient semantic information, while larger dimensions (e.g., 4096 or 6144) lead to sparse representations, thereby impeding model convergence. Se identified that an optimal embedding dimension of 2048 is suitable for images with an input resolution of 256 $\times$ 256 pixels.

\begin{figure}[!t]
    \centering
    \includegraphics[width=0.9\linewidth]{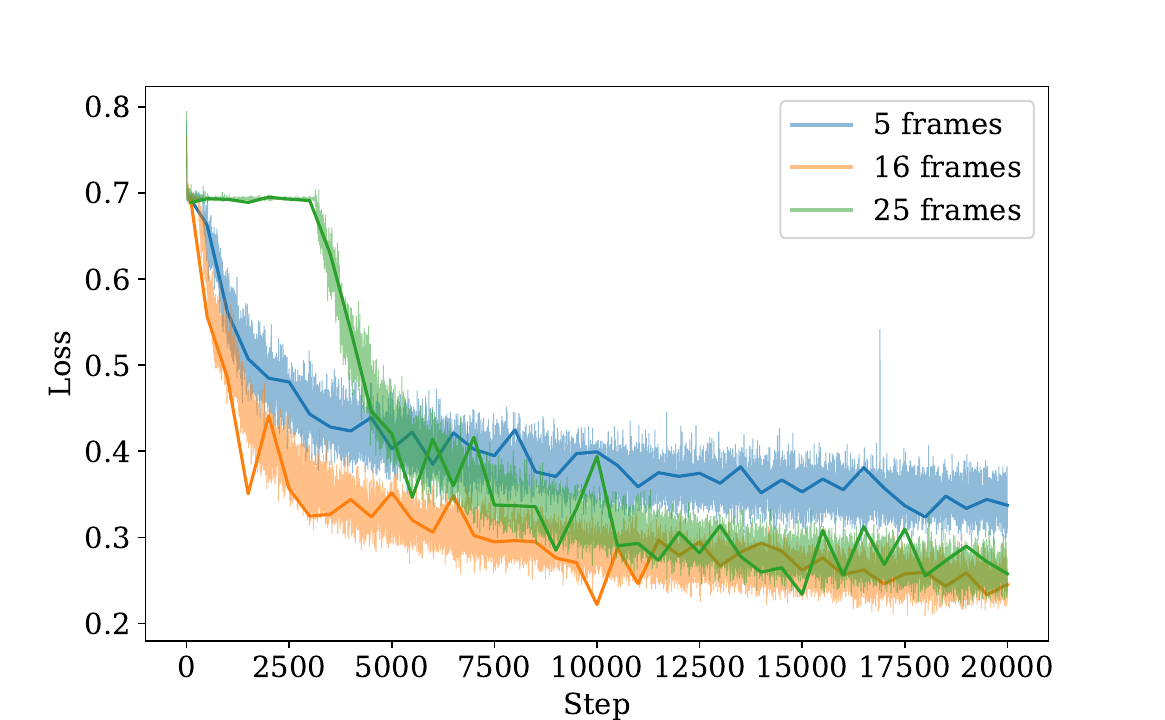}
    \caption{SyncNet training curves of different numbers of input frames. (VoxCeleb2, Batch size 512, StableSyncNet arch, Dim 2048.)}
    \vspace{-10pt}
    \label{fig:frames_ablation}
\end{figure}

\para{Number of frames.}
The number of input frames determines the range of visual and audio information that SyncNet can perceive. As shown in \cref{fig:frames_ablation}, selecting a larger number of frames (e.g., 16) can help the model converge. However, an excessively large number of frames (e.g., 25) will cause the model to get stuck around 0.69 in the early stage of training, and the validation loss at the end of training does not show a significant advantage compared to using 16 frames. 


\begin{figure}[!t]
    \centering
    \includegraphics[width=0.9\linewidth]{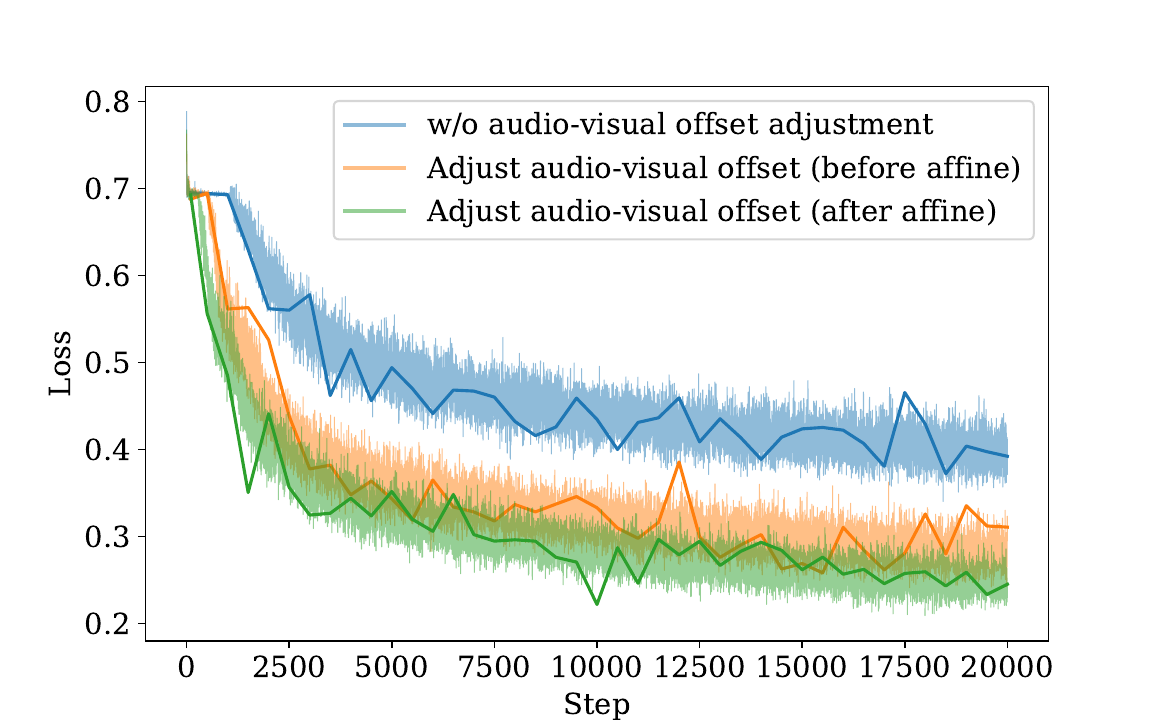}
    \caption{SyncNet training curves of different data preprocessing methods. (VoxCeleb2, Batch size 512, StableSyncNet arch, Dim 2048, 16 frames.)}
    \vspace{-10pt}
    \label{fig:offset_ablation}
\end{figure}

\para{Data preprocessing.}
In-the-wild videos naturally contain audio-visual offsets, it is necessary to adjust this offset to zero before inputting them into the SyncNet network. We used the official open-source version of pretrained SyncNet \cite{chung2017out} to adjust the offset and remove videos with ${\text{Sync}}_{\text{conf}}$ below 3. Specifically, we evaluated adjusting the offset before and after applying affine transformation. As shown in \cref{fig:offset_ablation}, without offset adjustment, the model's convergence is significantly impaired. Performing affine transformation before offset adjustment yields better results. This may be due to that affine transformation reducing data with side profiles or unusual angles, allowing the pretrained SyncNet \cite{chung2017out} to predict the offset more accurately.

\para{Discussions.}
Based on the ablation studies above, we found that batch size, number of input frames, and data preprocessing method are the primary factors affecting SyncNet convergence. We identified the optimal settings for a StableSyncNet with the 256 $\times$ 256 input size: batch size of 1024, 16 frames, SD U-Net encoder adapted for both visual and audio encoders, embedding dimension of 2048, and adjusting the audio-visual offset after affine transformation. We train a StableSyncNet on VoxCeleb2 \cite{chung2018voxceleb2} and test it on HDTF \cite{zhang2021flow}, which is an out-of-distribution experimental setup. The validation loss on VoxCeleb2 reaches around 0.18 and the accuracy on HDTF achieves 94\%, which significantly surpasses the previous SOTA result 91\% \cite{prajwal2020lip,gupta2023towards}.
\section{Experiments}

\begin{table*}[t]
    \centering
    \small
    \setlength{\cmidrulewidth}{0.01em}
    \renewcommand{\tabcolsep}{4pt}
    \renewcommand{\arraystretch}{1.1}
    \resizebox{0.8\linewidth}{!}{%
        \begin{tabular}{lcccccccccc}
            \hline
            \multirow{2}{*}{Method}                       & \multicolumn{4}{c}{HDTF} & \multicolumn{4}{c}{VoxCeleb2}                                                                                                                                                                                                        \\ \cmidrule[\cmidrulewidth](l){2-6} \cmidrule[\cmidrulewidth](l){7-11}
                                                          & FID $\downarrow$         & SSIM $\uparrow$               & ${\text{Sync}}_{\text{conf}}$ $\uparrow$ & LMD $\downarrow$ & FVD $\downarrow$ & FID $\downarrow$ & SSIM $\uparrow$ & ${\text{Sync}}_{\text{conf}}$ $\uparrow$ & LMD $\downarrow$ & FVD $\downarrow$ \\ \hline
            Wav2Lip \cite{prajwal2020lip}                 & 12.5                     & 0.70                          & 8.2                                      & 0.34             & 304.35           & 10.8             & 0.71            & 7.0                                      & 0.53             & 257.85           \\
            VideoReTalking \cite{cheng2022videoretalking} & 9.5                      & 0.75                          & 7.5                                      & 0.49             & 270.56           & 7.5              & 0.77            & 6.4                                      & 0.60             & 215.67           \\
            Diff2Lip \cite{mukhopadhyay2024diff2lip}      & 10.3                     & 0.72                          & 7.9                                      & 0.36             & 260.45           & 9.8              & 0.73            & 6.9                                      & 0.54             & 210.45           \\
            MuseTalk \cite{zhang2024musetalk}             & 9.35                     & 0.74                          & 6.8                                      & 0.56             & 246.75           & 7.1              & 0.80            & 5.9                                      & 0.64             & 203.43           \\ \hline
            LatentSync (Ours)                             & \textbf{7.22}            & \textbf{0.79}                 & \textbf{8.9}                             & \textbf{0.30}    & \textbf{162.74}  & \textbf{5.7}     & \textbf{0.81}   & \textbf{7.3}                             & \textbf{0.51}    & \textbf{123.27}  \\ \hline
        \end{tabular}%
    }
    \caption{Quantitative comparisons on HDTF and VoxCeleb2.}
    \vspace{-10pt}
    \label{table:quantitative_comparisons}
\end{table*}

\begin{figure*}[!t]
    \centering
    \includegraphics[width=\linewidth]{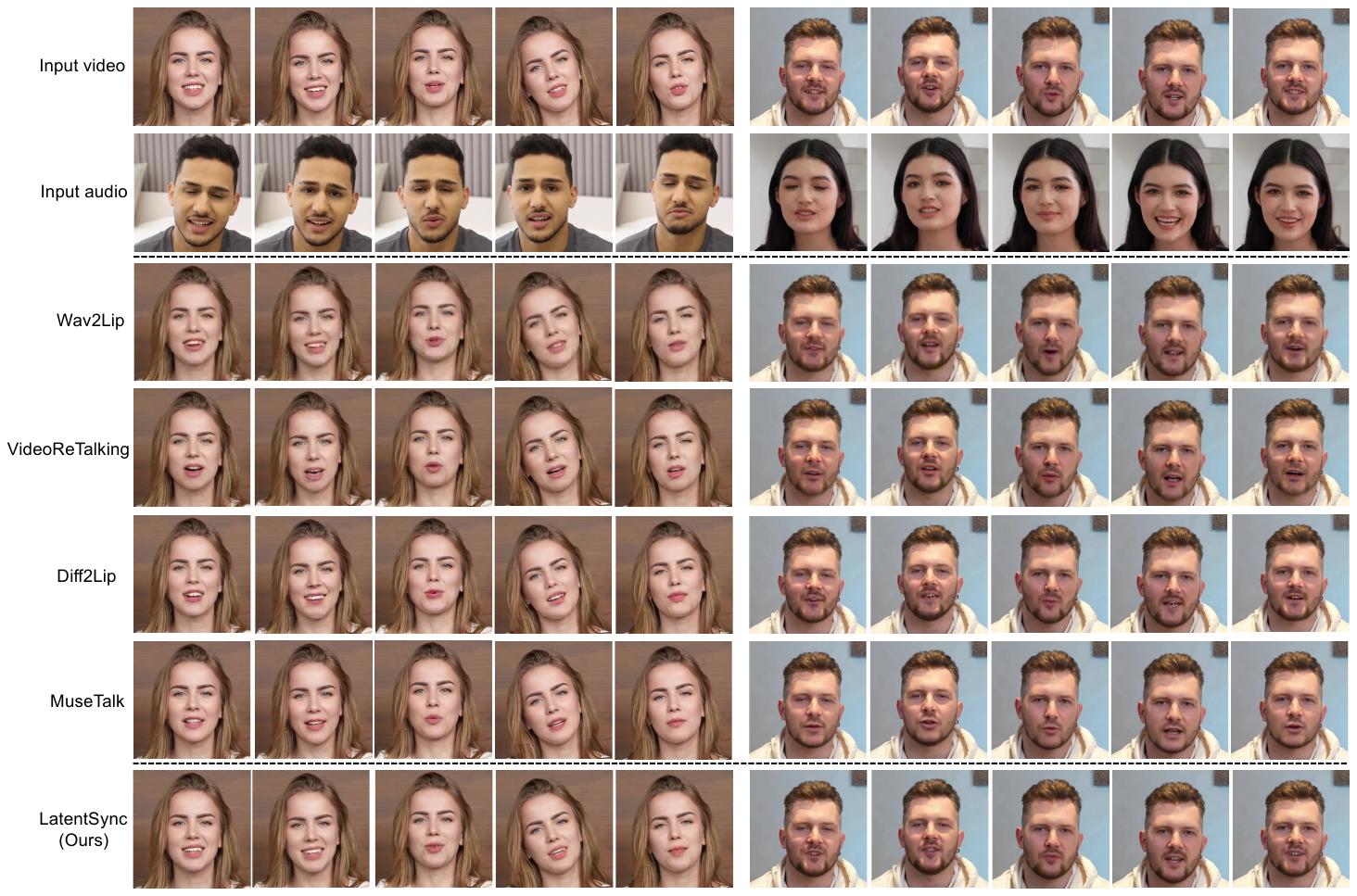}
    \caption{Qualitative comparisons with SOTA lip-sync methods. We run two cases in the cross generation setting \cite{mukhopadhyay2024diff2lip}. The first row demonstrates the original input video, and the second row is the video from which we extracted the audio as input, the video can be regarded as the target lip movements. Rows 3 $\sim$ 7 display the lip-synced videos. (All the photorealistic portrait images in this paper are from contracted models.)}
    \vspace{-10pt}
    \label{fig:comparisons}
\end{figure*}

\subsection{Experimental Settings}
\para{Datasets.}
We used a mixture of VoxCeleb2 \cite{chung2018voxceleb2} and HDTF \cite{zhang2021flow} datasets as our training set. VoxCeleb2 is a large-scale audio-visual dataset containing over 1 million utterances from over 6,000 speakers. It includes speakers from a wide range of ethnicities, accents, and backgrounds. HDTF contains 362 different high-definition (HD) videos, with resolutions typically around 720p to 1080p.

We used HyperIQA \cite{su2020blindly} to filter out videos with low visual quality, specifically blurry or pixelated videos. During evaluation, we randomly selected 30 videos from the test set of HDTF or VoxCeleb2.

\para{Implementation details.}
When evaluating our model LatentSync, we first converted the videos to 25 FPS, then applied the affine transformation based on facial landmarks detected by face-alignment \cite{bulat2017far} to obtain 256 $\times$ 256 face videos. The audio was resampled to 16kHz. We used 20 steps of DDIM \cite{song2020denoising} sampling for inference.

\para{Evaluation metrics.}
We evaluate our method in three aspects: (1) Visual quality. We use SSIM \cite{wang2004image} in the reconstruction setting and FID \cite{heusel2017gans} in the cross generation setting to assess visual quality. Following \cite{mukhopadhyay2024diff2lip}, the reconstruction setting refers to using the same audio as the input video, while the cross generation setting refers to using audio different from the input video. (2) Lip-sync accuracy. We use the confidence score of SyncNet (${\text{Sync}}_{\text{conf}}$) \cite{chung2017out} and the landmark distances around the mouth (LMD) \cite{guan2023stylesync}. We found that ${\text{Sync}}_{\text{conf}}$ aligns closely with visual assessments. (3) Temporal consistency. We adopt the widely used FVD metric \cite{unterthiner2018towards}.

\subsection{Comparisons}
\para{Comparison methods.}
We selected several SOTA methods that provide open-source inference code and checkpoints for comparison. Wav2Lip \cite{prajwal2020lip} is the classic lip-sync method, introducing the idea of using a pretrained SyncNet for supervision instead of a lip-sync discriminator \cite{kr2019towards}. VideoReTalking \cite{cheng2022videoretalking} divides the lip-sync process into three steps to improve the results. Diff2Lip \cite{mukhopadhyay2024diff2lip} utilized pixel-space diffusion models to achieve generalized lip sync. MuseTalk \cite{zhang2024musetalk} utilizes the architecture of Stable Diffusion for inpainting, but it is not based on the diffusion model framework and appears more like a GAN-based approach.

\para{Quantitative comparisons.}
As shown in \cref{table:quantitative_comparisons}, the lip-sync accuracy of our method significantly surpasses that of other methods. This is attributed to our 94\% accuracy StbaleSyncNet, as well as the audio cross-attention layers in U-Net, which better captures the relationship between audio and lip movements. In terms of visual quality, our method outperforms others, likely due to the powerful capabilities of the Stable Diffusion model. Owing to temporal layer \cite{guo2023animatediff} and the incorporation of TREPA, our FVD score is also superior to other methods.

\begin{table}[t]
    \centering
    \resizebox{0.9\columnwidth}{!}{
        \begin{tabular}{lcccc}
            \toprule
            Method                            & ${\text{Sync}}_{\text{conf}}$ $\uparrow$ & FVD $\downarrow$ \\
            \midrule
            LatentSync w/o SyncNet            & 4.6                                      & 220.37           \\
            LatentSync + latent space SyncNet & 7.9                                      & 180.45           \\
            LatentSync + pixel space SyncNet  & \textbf{8.9}                             & \textbf{162.74}  \\
            \bottomrule
        \end{tabular}
    }
    \caption{The ablation studies of different SyncNets. HDTF results.}
    \vspace{-10pt}
    \label{table:syncnet_ablation}
\end{table}

\para{Qualitative comparisons.}
We run two cases in the cross generation setting. According to \cref{fig:comparisons}, Wav2Lip has excellent lip-sync accuracy, but the generated videos are very blurry. VideoReTalking shows strange artifacts. Diff2Lip is limited by pixel-space diffusion models and can only generate low-resolution videos, resulting in noticeable blurriness. MuseTalk does not preserve facial features well, the man's beard becomes sparse. In contrast, our method excels in both clarity and identity preservation, even the mole on the woman's face was preserved. Furthermore, it hardly shows generated box due to the smooth shape of fixed mask.

\begin{figure}[!t]
    \centering
    \includegraphics[width=0.9\linewidth]{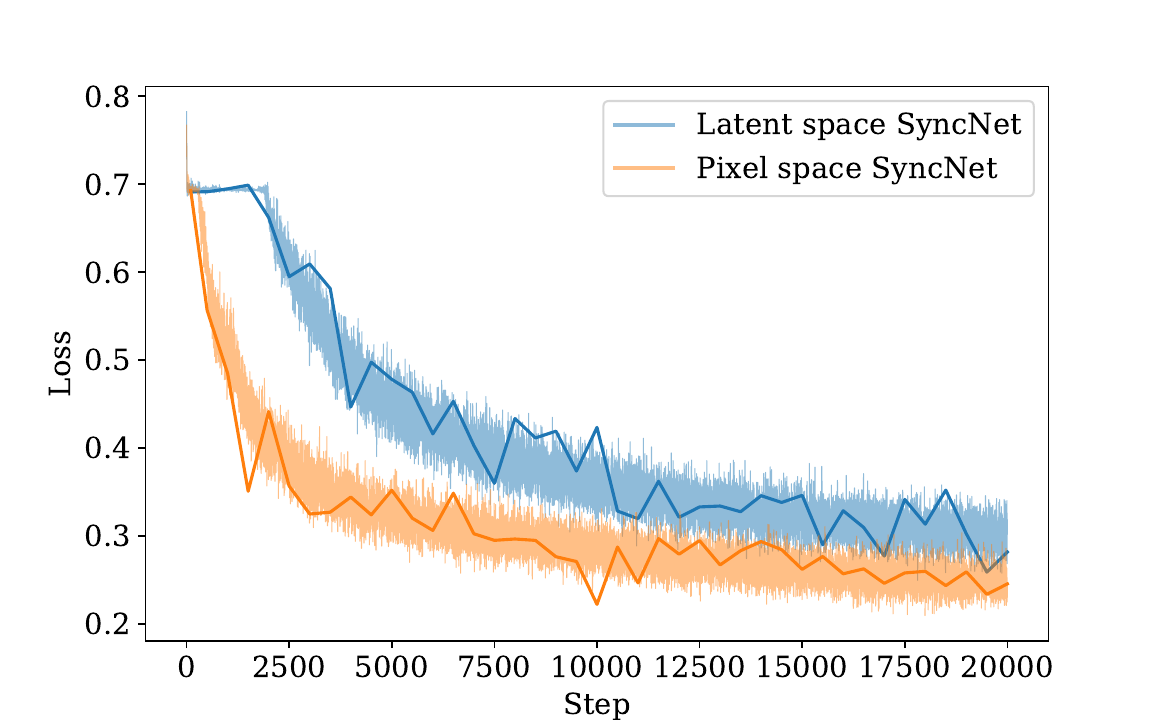}
    \caption{SyncNet training curves of different visual input spaces. Here we use the open-sourced pretrained VAE from Stability AI \cite{sd_vae_ft_mse} for encoding. (VoxCeleb2, Batch size 512, StableSyncNet arch, Dim 1024, 16 frames.)}
    \vspace{-10pt}
    \label{fig:space_ablation}
\end{figure}

\subsection{Ablation studies}
\label{sec:ablation}
\para{The effectiveness of SyncNet supervision.}
As shown in \cref{table:syncnet_ablation}, when the SyncNet supervision is not added, The lip-sync performance of the trained diffusion models is significantly poor. In fact, we found that other end-to-end lipsync methods \cite{prajwal2020lip,mukhopadhyay2024diff2lip} exhibit similar phenomena. As for supervision in two different spaces, diffusion models under pixel space SyncNet supervision perform better in terms of lip-sync accuracy and temporal consistency. This is likely due to the poor convergence of latent space SyncNet (as shown in \cref{fig:space_ablation}), which is reasonable since the input to the latent space SyncNet is the compressed latents obtained from VAE encoding, and some lip information may already be lost.

In addition, we found that improvements in lip-sync accuracy were accompanied by an increase in temporal consistency. This is because the audio window inherently encapsulates rich temporal information. Enhanced utilization of audio information by the model also leads to the improved temporal consistency.

\para{The effectiveness of TREPA.}
According to \cref{table:trepa_ablation}, both the temporal consistency and visual quality improve after incorporating TREPA. This improvement may be attributed to that the robust representations extracted by VideoMAE-v2 \cite{wang2023videomae} inherently encapsulate both visual and temporal information.

\begin{table}[t]
    \centering
    \resizebox{0.9\columnwidth}{!}{
        \begin{tabular}{lcccc}
            \toprule
            Method             & FID $\downarrow$ & SSIM $\uparrow$ & FVD $\downarrow$ \\
            \midrule
            LatentSync         & 7.71              & 0.77            & 176.35           \\
            LatentSync + TREPA & \textbf{7.22}     & \textbf{0.79}   & \textbf{162.74}  \\
            \bottomrule
        \end{tabular}
    }
    \caption{The ablation studies of TREPA. HDTF results.}
    \vspace{-10pt}
    \label{table:trepa_ablation}
\end{table}

\section{Conclusion}
We introduced LatentSync, the first lip-sync method based on audio-conditioned LDMs, which addresses the problems of traditional diffusion-based lip-sync methods: (1) The low sampling speed and inability to generate high-resolution videos of pixel space diffusion. (2) The information loss in two-stage generation methods. We identified the shortcut learning problem in lip-sync task and explored different approaches to incorporate SyncNet supervision to solve this problem. We conducted comprehensive ablation studies to find out several key factors influencing SyncNet converge. In addition, we proposed TREPA to further improve the temporal consistency of our method.

{
    \small
    \bibliographystyle{ieeenat_fullname}
    \bibliography{refs}
}

\clearpage
\onecolumn
\setcounter{page}{1}
\appendix
\begin{center}
    \LARGE\textbf{Appendix}
\end{center}


\section{SyncNet Training Loss Proof}
\label{suppsec:syncnet_loss_proof}

According to the classic training framework of SyncNet \cite{prajwal2020lip,chung2017out}, we randomly provide SyncNet with positive and negative samples with a 50\% probability during the training stage. The output of SyncNet is a probability distribution of whether the sample is positive or negative. We define $p(x=1)$ as the probability that the sample is positive, and $p(x=0)$ as the probability that the sample is negative. Let $p(x)$ be the true probability distribution and $q(x)$ the predicted probability distribution. We plot some scatter charts to observe the changes in the prediction probabilities of a non-converging SyncNet during the training process.
\begin{figure}[!htbp]
    \centering
    \includegraphics[width=\linewidth]{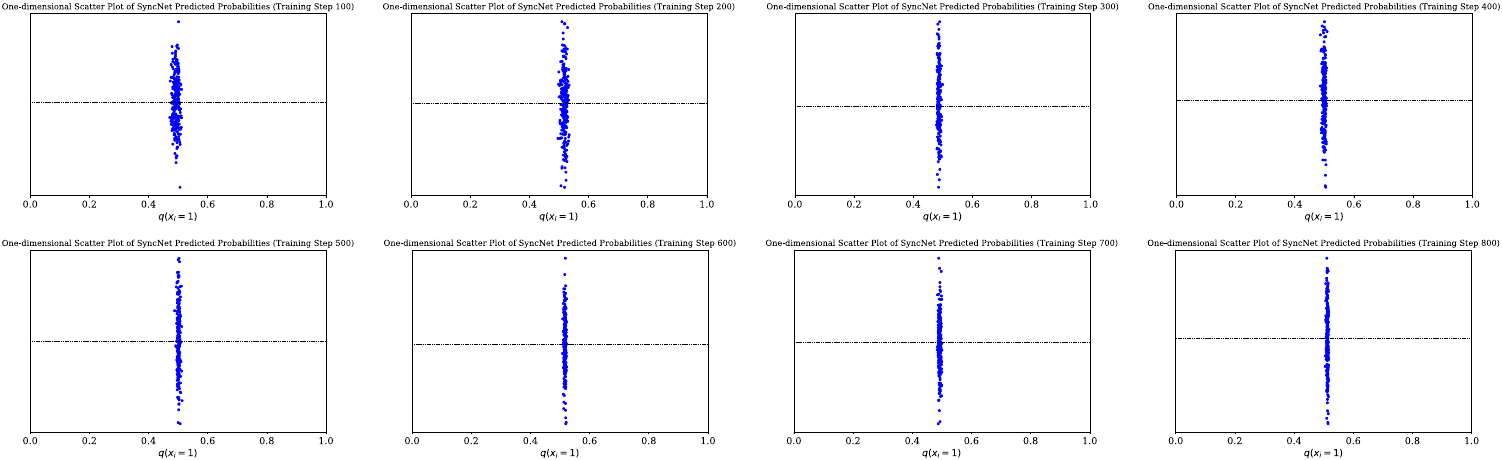}
    \caption{We train a SyncNet on VoxCeleb2 \cite{chung2018voxceleb2} with batch size of 256. We modify its architecture and data to make it non-converging. Every 100 steps, we plot a scatter plot to observe the probability distribution of SyncNet's predicted $q(x_i=1)$, including 256 data points. The x-axis represents the probability $q(x_i=1)$, and we add some random jitter along the y-axis for better visualization.}
    \label{fig:scatter}
\end{figure}

As shown in \cref{fig:scatter}, in the early stages of training, the probabilities predicted by SyncNet for $q(x_i=1)$ are dispersed, but later in the training, the probabilities for $q(x_i=1) $ are almost all distributed around 0.5 (the scatter charts for $q(x_i=0)$ are similar). Now we have:
\begin{align}
    \forall i \in {1, 2, \ldots, N}: ~~ q(x_i=1) \approx q(x_i=0) \approx 0.5
    \label{eq:q_prob}
\end{align}
We speculate that this may be because, the non-converging SyncNet is consistently unable to correctly distinguish whether a sample is positive or negative during the training process. However, to reduce the training loss, it tends to output a probability of 0.5 for all input embeddings to minimize the training loss as much as possible. This is why the training loss decreases somewhat initially and then gets stuck (from 0.8 to 0.69).

We assmue that there are $A$ positive samples and $B$ negative samples in a batch, and then we can derive SyncNet's training loss based on the following steps:
\begin{align}
    \mathcal{L}_{\text{syncnet}} & = -\frac{1}{N}\sum_{i=1}^N \sum_{x_i \in \{0,1\}} p(x_i) \,\text{log}\, {q(x_i)}                                                                                                                                                                                  \\
                                 & = -\frac{1}{N}\sum_{i=1}^N \left[ p(x_i=1) \,\text{log}\, {q(x_i=1)} + p(x_i=0) \,\text{log}\, {q(x_i=0)} \right]                                                                                                                                                 \\
                                 & = -\frac{1}{N}\sum_{i=1}^N p(x_i=1) \,\text{log}\, {q(x_i=1)} -\frac{1}{N}\sum_{i=1}^N p(x_i=0) \,\text{log}\, {q(x_i=0)}                                                                                                                                         \\
                                 & = -\frac{1}{N} \left[ \sum_{a=1}^A 1 \times \,\text{log}\, {q(x_a=1)} + \sum_{b=1}^B 0 \times \,\text{log}\, {q(x_b=1)} \right] -\frac{1}{N} \left[ \sum_{a=1}^A 0 \times \,\text{log}\, {q(x_a=1)} + \sum_{b=1}^B 1 \times \,\text{log}\, {q(x_b=1)} \right]     \\
                                 & = -\frac{1}{N} \left[ \sum_{a=1}^A \,\text{log}\, {q(x_a=1)} + \sum_{b=1}^B \,\text{log}\, {q(x_b=0)} \right]                                                                                                                                                     \\
                                 & \approx -\frac{1}{N} \left[ A \, \text{log}\, 0.5 + B \, \text{log}\, 0.5 \right]           \quad \quad \quad \quad \quad \quad \quad \quad \quad \quad \quad \quad \quad \quad \quad \quad \quad \quad \quad \quad \quad \quad     \text{Apply \cref{eq:q_prob}} \\
                                 & = 0.693
\end{align}

\end{document}